\documentclass[10pt]{article}
\usepackage{cite}
\usepackage{amsmath}
\usepackage{amssymb}
\usepackage{xcolor}
\usepackage{graphicx}
\usepackage{dblfloatfix}
\usepackage{multirow}
\usepackage{caption}
\usepackage{adjustbox}
\usepackage[margin=1in]{geometry}
\usepackage{lipsum}% http://ctan.org/pkg/lipsum

\title{\LARGE \bf
\emph{daVinciNet}: Joint Prediction of Motion and\\ Surgical State in Robot-Assisted Surgery}

\author{Yidan Qin$^{1,2}$, Seyedshams Feyzabadi$^{1}$, Max Allan$^{1}$, Joel W. Burdick$^{2}$, Mahdi Azizian$^{1}$% <-this % stops a space
\thanks{$^{1}$Intuitive Surgical Inc., 1020 Kifer Road, Sunnyvale, CA,94086, USA}%
\thanks{$^{2}$Department of Mechanical and Civil Engineering, California Institute of Technology, 1200 E California Blvd, Pasadena, CA, 91125, USA}%
\thanks{Emails: Ida.Qin@intusurg.com, Mahdi.Azizian@intusurg.com}
}

\makeatletter
\def\thanks#1{\protected@xdef\@thanks{\@thanks
        \protect\footnotetext{#1}}}
\makeatother

\date{\vspace{-5ex}}

\begin{document}

\maketitle
\thispagestyle{empty}
\pagestyle{empty}

%%%%%%%%%%%%%%%%%%%%%%%%%%%%%%%%%%%%%%%%%%%%%%%%%%%%%%%%%%%%%%%%%%%%%%%%%%%%%%%%
\begin{abstract}
\noindent
This paper presents a technique to concurrently and jointly predict the future trajectories of surgical instruments and the future state(s) of surgical subtasks in robot-assisted surgeries (RAS) using multiple input sources. Such predictions are a necessary first step towards shared control and supervised autonomy of surgical subtasks. Minute-long surgical subtasks, such as suturing or ultrasound scanning, often have distinguishable tool kinematics and visual features, and can be described as a series of fine-grained states with transition schematics. We propose \emph{daVinciNet} - an end-to-end dual-task model for robot motion and surgical state predictions. \emph{daVinciNet} performs concurrent end-effector trajectory and surgical state predictions using features extracted from multiple data streams, including robot kinematics, endoscopic vision, and system events. We evaluate our proposed model on an extended Robotic Intra-Operative Ultrasound (RIOUS+) imaging dataset collected on a da Vinci\textsuperscript{\textregistered} Xi surgical system and the JHU-ISI Gesture and Skill Assessment Working Set (JIGSAWS). Our model achieves up to 93.85\% short-term (0.5s) and 82.11\% long-term (2s) state prediction accuracy, as well as 1.07mm short-term and 5.62mm long-term trajectory prediction error. 
\end{abstract}

%%%%%%%%%%%%%%%%%%%%%%%%%%%%%%%%%%%%%%%%%%%%%%%%%%%%%%%%%%%%%%%%%%%%%%%%%%%%%%%%
\section{INTRODUCTION}

The implementation of autonomy in the field of surgical robotics, from passive functionalities such as virtual fixtures \cite{selvaggio2018passive} to autonomous surgical tasks\cite{shademan2016supervised,pedram2017autonomous}, has attracted the attention of many. Such systems enrich the manual teleoperation experience in robot-assisted surgeries (RAS) and assist the surgeons in many ways. Enhancements include automated changes in the user interface during surgery, additional surgeon-assisting system functionalities, and shared control or even autonomous tasks\cite{chalasani2018computational,dimaio2018interactive,kim2015automated}. In 2016, Yang et al. proposed a definition of the levels of autonomy in medical robotics, ranging from mechanical robot guidance to fully autonomous surgical procedures\cite{yang2017medical}, where the sensing of the user's desires play an integral role. One prerequisite for the applications mentioned above is the ability to anticipate the surgeon's intention and the robot's motions. Prediction of the robotic surgical instruments' trajectories, for instance, contributes to collision prediction and avoidance, including collisions between instruments or with obstacles in the proximity. It also has applications to safe multi-agent surgical systems where various surgical tasks are performed concurrently. Weede et al. presented an instrument trajectory prediction method for optimal endoscope positioning through autonomous endoscopic guidance \cite{weede2011intelligent}. The prediction of the next fine-grained surgical states, either fine-grained action states (picking up a needle)\cite{ahmidi2017dataset} or surgical phases (bladder dissection)\cite{zia2017temporal}, is useful in many surgeon-assisting features. Examples include the predictive triggering of cloud-based features or heavy-processing services which are inherently time consuming. This functionality provides a more seamless operational workflow. The prediction of the next surgical state also allows for more synchronized collaborations between the surgeons and operating room staff through workflow recognition \cite{padoy2019machine, staub2012human}.

\begin{figure}
    \centering
    \includegraphics[width=10cm]{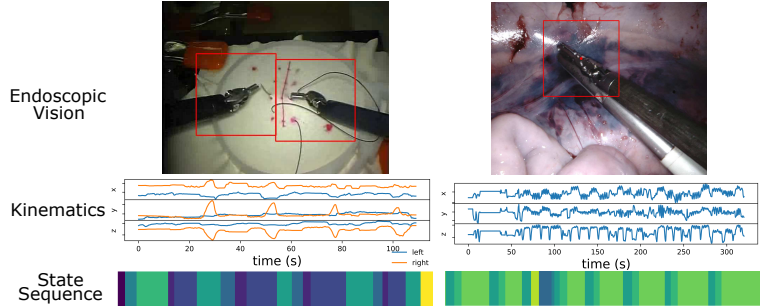}
    \caption{Sample data from JIGSAWS (left) and RIOUS+ (right). The top row shows the endoscopic vision, with RoI bounding boxes (areas surrounding the end-effectors). The middle row shows the end-effector paths in Cartesian coordinates. The bottom row shows the state sequences.}
    \label{fig:Figure1}
\end{figure}

Prediction of a robots surgical instruments' motion during a surgical task has found applications in surgical instrument tracking \cite{staub2010contour, zhao2012methods} and visual window tracking \cite{wang2018hybrid,sun2020visual}. Using only robot kinematics data, Staub et al. integrated surgical instrument trajectory and pose predictions with silhouette-based instrument tracking by using a Kalman filter to generate a pose prediction \cite{staub2010contour}. Similarly, instrument tracking procedures based on Kalman filtering with fusion of kinematics and visual data and visual matching of markers and appearance has been investigated \cite{zhao2012methods}. Additionally, a hybrid grey prediction model was implemented for autonomous endoscope navigation  during RAS \cite{wang2018hybrid}. Weede et al. proposed a guidance system for finding the optimal endoscope position through trajectory clustering and Markov Model (MM) for long-term instrument trajectory and optimal endoscope pose predictions\cite{weede2011intelligent}. These methods, however, all used surgical instrument motion prediction as auxiliary information to improve performance of their respective applications, without an extensive focus on improving motion prediction accuracy. Additionally, the prediction of surgical instrument trajectory seconds ahead has not been extensively researched. Weede et al. achieved long-term instrument trajectory prediction by trajectory clustering and aggregated time series data into categories \cite{weede2011intelligent}. Their MM-based model predicts the class of movements instead of the numerical temporal sequence of motion trajectory, which significantly limits its applications.

Surgical task such as suturing can be practically modeled as a Finite State Machine (FSM), with a list of discrete states (actions and non-actions) and possible transitions between states\cite{qin2020temporal}. Classically, the task is formulated as a MM and the transition probability matrix is learned from data \cite{tao2012sparse,rosen2006generalized, staub2012human, weede2011intelligent, volkov2017machine}. While powerful, Markov models do not capture temporal information, such as the duration of the last state and non-Markovian state sequences \cite{rosen2006generalized,weede2011intelligent}. With recent advances in machine learning, more learning-based state estimation methods have been proposed. Recurrent Neural Networks (RNN)\cite{dipietro2016recognizing} and Convolutional Neural Network (CNN) \cite{yu2018learning,lea2016segmental, lea2016temporal} models have achieved high levels of accuracy in state estimation. These models, however, perform state estimation independently without state transition information. Additionally, existing vision-based state estimation models have only extracted visual features from the entire endoscopic view \cite{lea2016temporal}. The emphasis on regions in the endoscopic view that are more indicative of surgical states (e.g., areas surrounding the end-effectors), is helpful for eliminating environmental noises in visual features. This can be done through instrument tracking \cite{allan20183}. 

Deep learning-based methods for path and action prediction have been used in the field of computer vision, including  path predictions using personal visual features and Long-Short Term Memory (LSTM) \cite{kooij2014context, yagi2018future} and early recognition of actions \cite{sadegh2017encouraging,ma2016learning}. But they have received little attention in surgical robotics. These methods have predicted human paths and actions seconds in advance. Liang et al. recently proposed a multi-task model for predicting a person's future path and activities in videos using various features, including the person's position, appearance, and interactions \cite{liang2019peeking}. Compared to human activity datasets (such as ActEV\cite{awad2018trecvid}) which are used for human path and activity predictions, our problem has the privilege of having synchronized robot kinematics, endoscopic vision, and system events as data sources. This is especially useful in the prediction of surgical states, since different sources of input data have their respective strengths and weaknesses in representing states with various kinematics and visual features. Previously, we have proposed a unified model for surgical state estimation - Fusion-KVE - that incorporated multiple types of input data and exceeded the state-of-the-art state estimation performance \cite{qin2020temporal}. Building on this, we explored the task of concurrent instrument path and surgical state predictions with multiple data streams and the incorporation of historic state transition sequences.

\textbf{Contributions}: This paper proposes \emph{daVinciNet}, a joint prediction model of instrument paths and surgical states for RAS tasks. Our model uses multiple data types gathered from a da Vinci\textsuperscript{\textregistered} surgical system as input (Fig. \ref{fig:Figure1}). The model performs feature extraction and makes multi-step predictions of both the end-effectors' trajectories in the endoscopic reference frame and the future surgical states. We aim for real-time predictions of up to 2 seconds in advance. Our main contributions include:

\vspace{5pt}
\begin{itemize}
    \item Incorporating and extracting features from multiple available data sources, including robot kinematics, endoscopic vision, and system events;
    \item Implementing a vision-based tool tracking algorithm to determine the Regions of Interest (RoIs) of the endoscopic vision data for more localized and effective visual feature extraction;
    \item Applying a state estimation model (Fusion-KVE) to infer the historic surgical state sequence for state prediction;
    \item Achieving accurate trajectory and state predictions for up to 2 seconds by incorporating the temporal information in data sequences using learning-based methods.
\end{itemize}
\vspace{5pt}

Our model's performance was evaluated using the JIGSAWS suturing dataset \cite{gao2014jhu} and the extended Robotic Intra-Operative Ultrasound (RIOUS+) imaging dataset \cite{qin2020temporal}. The RIOUS+ dataset contains ultrasound imaging trials in various experimental settings (phantom, in-vivo, and cadaver) and endoscopic motion. We performed multi-step end-effector path and surgical state predictions for various time spans (0.1s to 2s).  We also performed ablation studies \cite{meyes2019ablation} to better understand the contributions of various types of features used in both prediction tasks. To the best of our knowledge, there is no current benchmark for surgical instrument trajectory or state predictions. We show robustness of the \emph{daVinciNet} by comparing it to its ablated versions. We hope that this paper and the future release of the RIOUS+ dataset will encourage future exploration of surgical scene prediction. 

\section{Method}

\begin{figure*}
    \centering
    \medskip
    \includegraphics[width=\textwidth]{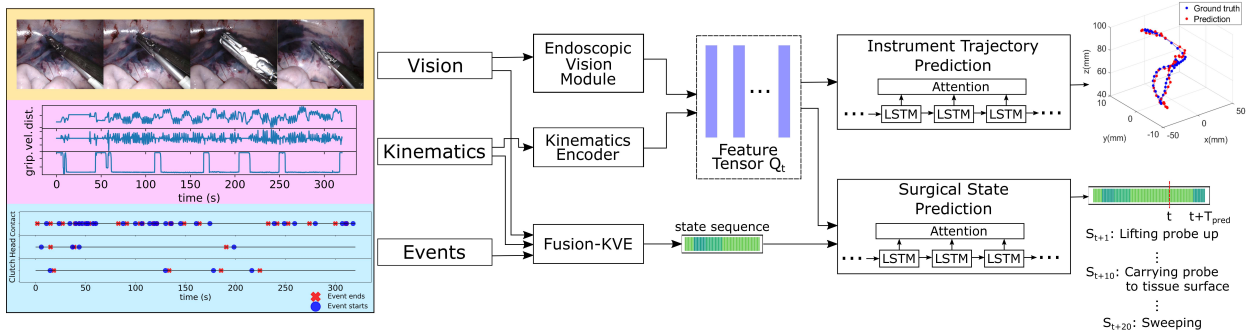}
    \caption{\emph{daVinciNet}'s model architecture. Given synchronized vision, robot kinematics and system events data streams, our model uses multiple encoders and Fusion-KVE to extract visual, kinematics, and states features. The concatenated feature tensor $\pmb{Q}$ is used for both instrument trajectory and surgical state predictions. The state sequence, in addition to $\pmb{Q}$, is the input of the surgical state prediction model. An attention-based LSTM decoder model was implemented for both prediction tasks. The example is shown with 10Hz data.}
    \label{fig:Figure2}
\end{figure*}

\begin{figure}
    \centering
    \includegraphics[width=10cm]{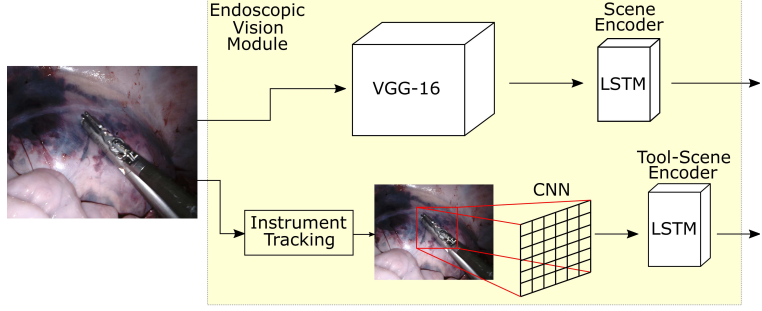}
    \caption{Details of the endoscopic vision module, which extracts both global visual information from the entire endoscopic view and RoI visual information that focuses on the interactions between the end-effector and the surrounding scene. The RoIs are determined by a silhouette-based instrument tracking model\cite{allan20183}.}
    \label{fig:Figure3}
\end{figure}

We propose an end-to-end joint prediction model that concurrently predicts the end-effector trajectories and the surgical states, as shown in Fig. \ref{fig:Figure2}. Our model resembles the structure of a Long-Short Term Memory (LSTM) encoder-decoder model - a model widely used in natural language processing \cite{cho2014properties}. It consists of feature extraction components that take in the vision, kinematics, and events data sequences for an observation window with size $T_{obs}$. The outputs of the endoscopic vision module and the kinematics encoder are consolidated to a feature tensor $\pmb{Q}$, which is fed to an attention-based LSTM model for multi-step trajectory prediction from time $t+1$ to $T_{pred}$, where $T_{pred}$ is the number of prediction steps. The output of Fusion-KVE is an input for the surgical state prediction in addition to the feature tensor. In the following subsections, we will first describe the details in feature extraction and prediction modules, and then the training details of the model. 

\subsection{Visual Feature Encoder}

We developed a novel endoscopic vision analysis module that extracts visual features at both global and local levels. The endoscopic scene features are extracted by a CNN-LSTM encoder model. We use a pre-trained VGG-16 model \cite{simonyan2014very} to extract a fixed size CNN feature vector from each frame in the endoscope video. Instead of directly using the temporal concatenation of CNN features as described in \cite{lea2016segmental,lea2016temporal,yu2018learning}, we drew inspiration from vision-based human path prediction \cite{alahi2016social,liang2019peeking} and implemented an LSTM encoder as the scene encoder (Fig. \ref{fig:Figure3}) to capture the the CNN feature series' long-term temporal dependencies. Given CNN features $\pmb{X}^{scene}_t=(\pmb{x}^{scene}_{t-T_{obs}+1}, ..., \pmb{x}^{scene}_t)$ with $\pmb{x}^{scene}_t \in \mathbb{R}^m$, where $m$ is the number of scene CNN features at each time-step, the LSTM encoder maps the hidden state $\pmb{h}^{scene}_t$ from $\pmb{x}^{scene}_t$ with:
  \begin{equation} \label{eq:LSTM_1}
  \pmb{h}^{scene}_t=LSTM(\pmb{h}^{scene}_{t-1}, \pmb{x}_t),
  \end{equation} 
where $\pmb{h}_t \in \mathbb{R}^{n_{scene}}$ and $n_{scene}$ is the LSTM encoder's hidden size. The concatenated encoder hidden states $\pmb{H}^{scene}_t=(\pmb{h}^{scene}_{t-T_{obs}+1}, ..., \pmb{h}^{scene}_t))$ forms a part of the feature tensor $\pmb{Q}_t$. 

The background environment in real-world RAS endoscopic vision is complex and varies significantly across cases. The effect of such environmental noise can be eliminated by using large annotated datasets with various backgrounds to train the aforementioned CNN-LSTM encoder; however, such datasets are expensive to acquire. It is worth noting that most states in surgical task FSMs are associated with the movements of the instrument end-effectors (Table I). We implemented a silhouette-based instrument tracking model following \cite{allan20183} and performed bounding box detection of the end-effectors from the endoscopic view. The surrounding areas (bounding box) of each instrument's coordinates from the endoscopic view are the RoIs, and were the input to a different CNN-LSTM encoder model for feature extraction (the tool-scene encoder in Fig. \ref{fig:Figure3}). The encoded RoI hidden state $\pmb{H}^{RoI}_t \in \mathbb{R}^{T_{obs} \times n_{RoI}}$ is also a part of $\pmb{Q}_t$.

\textbf{Implementation details}: Each endoscope video frame was resized to a $224 \times 224 \times 3$ RGB image before being input to the VGG-16 model. The VGG-16 model was pre-trained following our previous work \cite{qin2020temporal}. $m=1024$ CNN features were extracted. The original $1280 \times 1024 \times 3$ RGB image of each frame was then input to the instrument tracking model. If there are instruments in the endoscopic view, the RoIs around the instruments' end-effectors were extracted for 100 CNN features through two layers of CNNs with ReLU \cite{nair2010rectified} activation. The scene encoder and the tool-scene encoder both have $n_{scene}=n_{RoI}=32$ hidden states. 

\subsection{Attention-based Kinematics Feature Encoder}

To extract kinematics features from multiple da Vinci\textsuperscript{\textregistered} surgical system data inputs (end-effectors' translational and rotational positions, etc.), and capture the long-term data progress, we followed \cite{qin2017dual} and implemented an LSTM encoder with input attention to identify the importance of different driving series. At time $t$, the kinematics input is $\pmb{X}^{kin}_t=(\pmb{x}^{kin}_{t-T_{obs}+1}, ..., \pmb{x}^{kin}_t)$, where $\pmb{x}^{kin}_t \in \mathbb{R}^l$ and $l$ is the number of kinematics input series. Instead of deriving $\pmb{h}^{kin}_t$ directly from Eq. (\ref{eq:LSTM_1}), we constructed the input attention mechanism by learning a multiplier vector that represents the weights of each input series at time $t$ from the previous hidden state $\pmb{h}^{kin}_{t-1}$ and the LSTM unit's cell state $\pmb{s}^{kin}_{t-1}$:

\begin{equation}
    \pmb{\alpha}^i_t=softmax(u_e^T tanh(\pmb{W}_e(\pmb{h}^{kin}_{t-1}, \pmb{s}^{kin}_{t-1}) + \pmb{V}_e\pmb{x}^{kin,i})),
\end{equation} 
where $\pmb{x}^{kin,i} \in \mathbb{R}^{T_{obs}}$ is the i-th kinematics input series ($1 \leq i \leq l$), and $\pmb{W}_e$ and $\pmb{V}_e$ are learnable encoder parameters. A softmax function normalizes the attention weights $\pmb{\alpha}_t$. The weighted kinematics input at time $t$ is then:

\begin{equation}
    \tilde{\pmb{x}}^{kin}_t=\sum_{i=1}^m \alpha^i_t \pmb{x}^{kin,i}_t,
\end{equation} 
which substitutes $\pmb{x}_t$ in Eq. (\ref{eq:LSTM_1}). The encoded hidden states $\pmb{H}^{kin}_t=(\pmb{h}^{kin}_{t-T_{obs}+1}, ..., \pmb{h}^{kin}_t) \in \mathbb{R}^{T_{obs} \times n_{kin}}$ is the final component of feature tensor $\pmb{Q}_t$.

\textbf{Implementation details}: the JIGSAWS suturing dataset and the RIOUS+ dataset respectively contain 26 and 16 kinematic variables. Except for the target variables (the end-effectors' paths), we used $l=20, 13$ input series for the JIGSAWS and the RIOUS+ dataset, respectively. The kinematics encoder had $n_{kin}=32$ hidden states.

\subsection{Instrument Path and Surgical State Predictions}

After encoding, a feature tensor $\pmb{Q}=(\pmb{q}_{t-T_{obs}+1}, ..., \pmb{q}_t) \in \mathbb{R}^{T_{obs} \times (n_{scene}+n_{RoI}+n_{kin})}$ was obtained. We implemented LSTM decoders to predict the Cartesian instrument paths in the endoscopic reference frame after time $t$ ($\pmb{y}_t \in \mathbb{R}^{T_{pred} \times r}$) and future states ($s_t \in \mathbb{R}^{T_{pred}}$), respectively. $T_{pred}$ is the number of prediction time-steps. The LSTM decoders were implemented with temporal attention \cite{bahdanau2014neural} to alleviate the performance deterioration as the input sequences' lengths $T_{obs}$ increase \cite{cho2014properties}. The temporal attention mechanism allows the decoders to use relevant hidden states among all time-steps from $\pmb{Q}$ in an adaptive manner. At time $t$, the temporal attention weights $\pmb{\beta} \in \mathbb{R}^{T_{obs}}$ of the decoder hidden state $\pmb{d}_t \in \mathbb{R}^{n'}$ is learned from $\pmb{d}_{t-1}$, the previous cell state of the decoder LSTM unit $\pmb{c}_{t-1}$, and the feature tensor:
\begin{equation}
\label{eq:beta}
    \pmb{\beta}^j_t=softmax(u_d^T tanh(\pmb{W}_d(\pmb{d}_{t-1}, \pmb{c}_{t-1}) + \pmb{V}_d\pmb{q}_j)),
\end{equation} 
where $n'$ is the hidden size, $\pmb{W}_d$ and $\pmb{V}_d$ are decoder parameters to be learned. 

The weighted feature $\tilde{\pmb{q}}_t=\sum_{j=1}^{T_{obs}} \beta^j_t \pmb{q}^j_t$ and the historic target sequences (3-D end-effector path $\pmb{y}$ or estimated surgical state $s$) from $t-T_{obs}+1$ to $t$ were used to extract the target embedding following \cite{liang2019peeking}:
\begin{equation}
\label{eq:y_tilde}
    \tilde{\pmb{y}}_{t-1}=\pmb{W}_{targ}(\tilde{\pmb{q}}_{t-1}, \pmb{y}_{t-1})+\pmb{V}_{targ},
\end{equation} 
where $\pmb{W}_{targ}$ and $\pmb{V}_{targ}$ are learned. The update of $\pmb{d}_t$ is:
\begin{equation}
\label{eq:d_t}
    \pmb{d}_t=LSTM(\pmb{d}_{t-1},[\tilde{\pmb{y}}_{t-1},\tilde{\pmb{q}}_t]),
\end{equation} 
after which the end-effector trajectory predictions $\hat{\pmb{y}}_t$ are computed by a fully connected layer. The probability vector $\pmb{s}$ for surgical state prediction can be similarly derived. The state prediction $\hat{s}_t \in \mathbb{R}^{T_{pred}}$ is the future state sequence, with each state having the maximum likelihood among all states at each time-step.

It is worth noting that in order to obtain the historic sequence of surgical state $s$ from $t-T_{obs}+1$ to $t$, we implemented Fusion-KVE - a unified surgical state estimation model we proposed recently \cite{qin2020temporal}- instead of using the ground truth (GT) state sequence. In real-time RAS settings, the surgical state prediction model does not have access to the manually-labeled historic surgical state sequence; therefore, a state estimation model is needed to provide the historic state sequence. For the JIGSAWS suturing dataset, we implemented the ablated version of the state estimation model (Fusion-KV) due to the lack of system events data. Section \ref{sec:results} will discuss the performance difference when various features are included for both prediction tasks. 

\textbf{Implementation details}: We implemented both trajectory and state prediction LSTM decoders with $n'=96$ hidden states after grid search for parameters. $r=6$ variables (3D end-effector paths for both instruments) were predicted for the JIGSAWS data set, while $r=3$ variables were predicted for the RIOUS+ dataset.  Multi-step predictions were implemented, with $T_{obs}=20$ and $max(T_{pred})=20$ for data streaming at 10Hz. 

\subsection{Training}

The entire model, including the feature extraction and the prediction modules, was trained end-to-end with the goal of minimizing a loss function that accounts for both the trajectory prediction and state prediction accuracies. The trajectory loss function is the cumulative $L_2$ loss between the predicted end-effector trajectory and the GT trajectory, summed up from $T_{obs}+1$ to $T_{pred}$. The state estimation loss function is the cumulative categorical cross-entropy loss that accounts for the discrepancies between the predicted surgical states and the GT. 

\section{Experimental Evaluations}

\begin{table}[!b]
\footnotesize
\centering
\caption{Datasets State Descriptions and Duration}
\begin{tabular}{ccc}
\multicolumn{3}{c}{\textbf{JIGSAWS Suturing Dataset}}                                                         \\ \hline
\multicolumn{1}{c|}{Action ID} & \multicolumn{1}{c|}{Description}                              & Duration (s) \\ \hline
\multicolumn{1}{c|}{G1}        & \multicolumn{1}{c|}{Reaching for the needle with right hand}  & 2.2          \\
\multicolumn{1}{c|}{G2}        & \multicolumn{1}{c|}{Positioning the tip of the needle}        & 3.4          \\
\multicolumn{1}{c|}{G3}        & \multicolumn{1}{c|}{Pushing needle through the tissue}        & 9.0          \\
\multicolumn{1}{c|}{G4}        & \multicolumn{1}{c|}{Transferring needle from left to right}   & 4.5          \\
\multicolumn{1}{c|}{G5}        & \multicolumn{1}{c|}{Moving to center with needle in grip}     & 3.0          \\
\multicolumn{1}{c|}{G6}        & \multicolumn{1}{c|}{Pulling suture with left hand}            & 4.8          \\
\multicolumn{1}{c|}{G7}        & \multicolumn{1}{c|}{Orienting needle}                         & 7.7          \\
\multicolumn{1}{c|}{G8}        & \multicolumn{1}{c|}{Using right hand to help tighten suture}  & 3.1          \\
\multicolumn{1}{c|}{G9}        & \multicolumn{1}{c|}{Dropping suture and moving to end points} & 7.3          \\
\multicolumn{3}{c}{\textbf{RIOUS+ Dataset}}                                                      \\ \hline
\multicolumn{1}{c|}{State ID}  & \multicolumn{1}{c|}{Description}                              & Duration (s) \\ \hline
\multicolumn{1}{c|}{S1}        & \multicolumn{1}{c|}{Probe released, out of endoscopic view}   & 6.3         \\
\multicolumn{1}{c|}{S2}        & \multicolumn{1}{c|}{Probe released, in endoscopic view}       & 7.6         \\
\multicolumn{1}{c|}{S3}        & \multicolumn{1}{c|}{Reaching for probe}                       & 3.1          \\
\multicolumn{1}{c|}{S4}        & \multicolumn{1}{c|}{Grasping probe}                           & 1.1          \\
\multicolumn{1}{c|}{S5}        & \multicolumn{1}{c|}{Lifting probe up}                         & 2.4          \\
\multicolumn{1}{c|}{S6}        & \multicolumn{1}{c|}{Carrying probe to tissue surface}         & 2.3          \\
\multicolumn{1}{c|}{S7}        & \multicolumn{1}{c|}{Sweeping}                                 & 5.1          \\
\multicolumn{1}{c|}{S8}        & \multicolumn{1}{c|}{Releasing probe}                          & 1.7         
\end{tabular}
\end{table}

We evaluated our trajectory and state prediction models on the JIGSAWS and RIOUS+ data sets (Table I).

\begin{table}[!b]
\begin{adjustbox}{center}
\begin{tabular}{ccccccccccccccccc}
\multicolumn{10}{c}{\textbf{JIGSAWS Suturing}}                                                                                                                                                                                                                                                                                                                                              &                       & \multicolumn{6}{c}{\textbf{RIOUS+}}                                                                                                                                                                                                                                 \\ \cline{1-10} \cline{12-17} 
\multicolumn{1}{|c}{}                                                                                                   & \multicolumn{1}{c|}{}     & \multicolumn{1}{c|}{$x_1$} & \multicolumn{1}{c|}{$y_1$} & \multicolumn{1}{c|}{$z_1$} & \multicolumn{1}{c|}{$d_1$} & \multicolumn{1}{c|}{$x_2$} & \multicolumn{1}{c|}{$y_2$} & \multicolumn{1}{c|}{$z_2$} & \multicolumn{1}{c|}{$d_2$} & \multicolumn{1}{c|}{} &                                                                                                                        & \multicolumn{1}{c|}{}     & \multicolumn{1}{c|}{$x$}  & \multicolumn{1}{c|}{$y$}  & \multicolumn{1}{c|}{$z$}  & \multicolumn{1}{c|}{$d$}  \\ \cline{1-10} \cline{12-17} 
\multicolumn{1}{|c|}{\multirow{3}{*}{$\pmb{H}^{kin}$}}                                                                  & \multicolumn{1}{c|}{RMSE} & \multicolumn{1}{c|}{2.81}  & \multicolumn{1}{c|}{2.42}  & \multicolumn{1}{c|}{3.28}  & \multicolumn{1}{c|}{4.16}  & \multicolumn{1}{c|}{3.8}   & \multicolumn{1}{c|}{4.26}  & \multicolumn{1}{c|}{4.75}  & \multicolumn{1}{c|}{5.92}  & \multicolumn{1}{c|}{} & \multicolumn{1}{c|}{\multirow{3}{*}{$\pmb{H}^{kin}$}}                                                                  & \multicolumn{1}{c|}{RMSE} & \multicolumn{1}{c|}{1.67} & \multicolumn{1}{c|}{1.8}  & \multicolumn{1}{c|}{1.22} & \multicolumn{1}{c|}{2.3}  \\ \cline{2-10} \cline{13-17} 
\multicolumn{1}{|c|}{}                                                                                                  & \multicolumn{1}{c|}{MAE}  & \multicolumn{1}{c|}{2.19}  & \multicolumn{1}{c|}{1.95}  & \multicolumn{1}{c|}{2.86}  & \multicolumn{1}{c|}{3.7}   & \multicolumn{1}{c|}{3.42}  & \multicolumn{1}{c|}{3.91}  & \multicolumn{1}{c|}{4.31}  & \multicolumn{1}{c|}{5.34}  & \multicolumn{1}{c|}{} & \multicolumn{1}{c|}{}                                                                                                  & \multicolumn{1}{c|}{MAE}  & \multicolumn{1}{c|}{1.45} & \multicolumn{1}{c|}{1.62} & \multicolumn{1}{c|}{1.24} & \multicolumn{1}{c|}{2.06} \\ \cline{2-10} \cline{13-17} 
\multicolumn{1}{|c|}{}                                                                                                  & \multicolumn{1}{c|}{MAPE} & \multicolumn{1}{c|}{6.8}   & \multicolumn{1}{c|}{6.09}  & \multicolumn{1}{c|}{7.39}  & \multicolumn{1}{c|}{8.93}  & \multicolumn{1}{c|}{7.77}  & \multicolumn{1}{c|}{8.03}  & \multicolumn{1}{c|}{8.2}   & \multicolumn{1}{c|}{10.14} & \multicolumn{1}{c|}{} & \multicolumn{1}{c|}{}                                                                                                  & \multicolumn{1}{c|}{MAPE} & \multicolumn{1}{c|}{1.89} & \multicolumn{1}{c|}{2.62} & \multicolumn{1}{c|}{1.76} & \multicolumn{1}{c|}{2.17} \\ \cline{1-10} \cline{12-17} 
\multicolumn{1}{|c|}{\multirow{3}{*}{\begin{tabular}[c]{@{}c@{}}\{$\pmb{H}^{scene}$,\\ $\pmb{H}^{kin}$\}\end{tabular}}} & \multicolumn{1}{c|}{RMSE} & \multicolumn{1}{c|}{2.7}   & \multicolumn{1}{c|}{2.29}  & \multicolumn{1}{c|}{3.25}  & \multicolumn{1}{c|}{4.01}  & \multicolumn{1}{c|}{3.65}  & \multicolumn{1}{c|}{4.01}  & \multicolumn{1}{c|}{4.63}  & \multicolumn{1}{c|}{5.2}   & \multicolumn{1}{c|}{} & \multicolumn{1}{c|}{\multirow{3}{*}{\begin{tabular}[c]{@{}c@{}}\{$\pmb{H}^{scene}$,\\ $\pmb{H}^{kin}$\}\end{tabular}}} & \multicolumn{1}{c|}{RMSE} & \multicolumn{1}{c|}{1.67} & \multicolumn{1}{c|}{1.7}  & \multicolumn{1}{c|}{1.18} & \multicolumn{1}{c|}{2.1}  \\ \cline{2-10} \cline{13-17} 
\multicolumn{1}{|c|}{}                                                                                                  & \multicolumn{1}{c|}{MAE}  & \multicolumn{1}{c|}{2.17}  & \multicolumn{1}{c|}{1.88}  & \multicolumn{1}{c|}{2.79}  & \multicolumn{1}{c|}{3.5}   & \multicolumn{1}{c|}{3.15}  & \multicolumn{1}{c|}{3.7}   & \multicolumn{1}{c|}{4.16}  & \multicolumn{1}{c|}{4.76}  & \multicolumn{1}{c|}{} & \multicolumn{1}{c|}{}                                                                                                  & \multicolumn{1}{c|}{MAE}  & \multicolumn{1}{c|}{1.33} & \multicolumn{1}{c|}{1.52} & \multicolumn{1}{c|}{1.12} & \multicolumn{1}{c|}{1.91} \\ \cline{2-10} \cline{13-17} 
\multicolumn{1}{|c|}{}                                                                                                  & \multicolumn{1}{c|}{MAPE} & \multicolumn{1}{c|}{6.73}  & \multicolumn{1}{c|}{5.88}  & \multicolumn{1}{c|}{7.18}  & \multicolumn{1}{c|}{8.44}  & \multicolumn{1}{c|}{7.05}  & \multicolumn{1}{c|}{7.5}   & \multicolumn{1}{c|}{7.91}  & \multicolumn{1}{c|}{9.27}  & \multicolumn{1}{c|}{} & \multicolumn{1}{c|}{}                                                                                                  & \multicolumn{1}{c|}{MAPE} & \multicolumn{1}{c|}{1.7}  & \multicolumn{1}{c|}{2.43} & \multicolumn{1}{c|}{1.57} & \multicolumn{1}{c|}{2.01} \\ \cline{1-10} \cline{12-17} 
\multicolumn{1}{|c|}{\multirow{3}{*}{$\pmb{Q}$}}                                                                        & \multicolumn{1}{c|}{RMSE} & \multicolumn{1}{c|}{2.53}  & \multicolumn{1}{c|}{1.89}  & \multicolumn{1}{c|}{2.96}  & \multicolumn{1}{c|}{3.35}  & \multicolumn{1}{c|}{3.15}  & \multicolumn{1}{c|}{3.5}   & \multicolumn{1}{c|}{3.91}  & \multicolumn{1}{c|}{4.51}  & \multicolumn{1}{c|}{} & \multicolumn{1}{c|}{\multirow{3}{*}{$\pmb{Q}$}}                                                                        & \multicolumn{1}{c|}{RMSE} & \multicolumn{1}{c|}{1.23} & \multicolumn{1}{c|}{1.41} & \multicolumn{1}{c|}{1.08} & \multicolumn{1}{c|}{1.98} \\ \cline{2-10} \cline{13-17} 
\multicolumn{1}{|c|}{}                                                                                                  & \multicolumn{1}{c|}{MAE}  & \multicolumn{1}{c|}{2.07}  & \multicolumn{1}{c|}{1.51}  & \multicolumn{1}{c|}{2.46}  & \multicolumn{1}{c|}{3.09}  & \multicolumn{1}{c|}{2.78}  & \multicolumn{1}{c|}{3.06}  & \multicolumn{1}{c|}{3.5}   & \multicolumn{1}{c|}{4.17}  & \multicolumn{1}{c|}{} & \multicolumn{1}{c|}{}                                                                                                  & \multicolumn{1}{c|}{MAE}  & \multicolumn{1}{c|}{1.09} & \multicolumn{1}{c|}{1.34} & \multicolumn{1}{c|}{0.97} & \multicolumn{1}{c|}{1.64} \\ \cline{2-10} \cline{13-17} 
\multicolumn{1}{|c|}{}                                                                                                  & \multicolumn{1}{c|}{MAPE} & \multicolumn{1}{c|}{6.43}  & \multicolumn{1}{c|}{4.72}  & \multicolumn{1}{c|}{6.35}  & \multicolumn{1}{c|}{7.46}  & \multicolumn{1}{c|}{6.13}  & \multicolumn{1}{c|}{6.11}  & \multicolumn{1}{c|}{6.67}  & \multicolumn{1}{c|}{7.95}  & \multicolumn{1}{c|}{} & \multicolumn{1}{c|}{}                                                                                                  & \multicolumn{1}{c|}{MAPE} & \multicolumn{1}{c|}{1.31} & \multicolumn{1}{c|}{2.16} & \multicolumn{1}{c|}{1.1}  & \multicolumn{1}{c|}{1.72} \\ \cline{1-10} \cline{12-17} 
\end{tabular}
\end{adjustbox}
\caption*{Table II: End-effector trajectory prediction performance measures when predicting one second ahead ($T_{pred}=10$). The prediction performances for the Cartesian end-effector path in the endoscopic reference frame $(x,y,z)$ and $d=\sqrt{x^2+y^2+z^2}$ are compared when the trajectory prediction decoder uses only kinematics features ($\pmb{H}^{kin}$), uses global scene and kinematics features (\{$\pmb{H}^{scene}$, $\pmb{H}^{kin}$\}), and uses global scene, RoI, and kinematics features ($\pmb{Q}$).}
\end{table}

\subsection{Datasets}

\textbf{JIGSAWS}: The JIGSAWS dataset includes three types of RAS tasks performed in a benchtop setting \cite{ahmidi2017dataset}. Each trial lasts around 1.5 minutes and contains synchronized endoscopic video and robot kinematics data. We used the 39 suturing trials for model evaluation, which has 9 possible actions (Table I). The kinematics data series included the end-effectors' positions, velocities, and gripper angles of the universal patient-side manipulators (USM). We converted the rotation matrices that represent the two end-effectors' orientations into Euler angles to reduce data dimensionality.

\textbf{RIOUS+}: The RIOUS+ dataset is an extended version of RIOUS, which was first introduced in \cite{qin2020temporal}, where the dataset was used to evaluate surgical state estimation accuracy. The surgical task performed is ultrasound scanning, which is a commonly used da Vinci intra-operative procedure to understand a patient's anatomic structures. Comparing to JIGSAWS, the RIOUS+ dataset contains more real-world RAS elements. RIOUS+ contains 40 trials performed by 5 users, 27 of which were performed on a phantom kidney in a dry-lab setting, 9 were on a porcine kidney, and 4 were on a cadaver liver performed in operating room settings. For each trial, RIOUS+ includes synchronized endoscopic vision, robot kinematics, and system events data collected from a da Vinci\textsuperscript{\textregistered} Xi surgical system. The position, velocity, and gripper angles of the USM were included as well as the endoscope position. The system events data are represented as binary time series, including surgeon head in/out, two ultrasound probe events, master clutch, camera follow, and instrument follow. The ultrasound imaging task has 8 possible states (Table I). 

\subsection{Metrics}

We use three metrics to evaluate the accuracy of our end-effector trajectory prediction: Root Mean Squared Error (RMSE), Mean Absolute Error (MAE), and Mean Absolute Percentage Error (MAPE) \cite{qin2017dual,plutowski1996experience}:
\begin{equation}
\begin{split}
& RMSE=\sqrt[]{\frac{\sum_{i=1}^N (y^i-\hat{y}^i)^2}{N}} \\
& MAE=\frac{\sum_{i=1}^N |y^i-\hat{y}^i|}{N} \\
& MAPE=\sum_{i=1}^N \left| \frac{y^i-\hat{y}^i}{y^i} \right| \times 100\% 
\end{split}
\end{equation}
Since RMSE and MAE are independent of the variables' absolute values, they provide an intuitive comparison among variables in the same datset. MAPE calculates the percentage error; therefore, it provides a direct comparison between prediction accuracies across different datasets. To evaluate the surgical state prediction accuracy, we calculated the percentage of accurately predicted frames in the testing sequences \cite{lea2016temporal,qin2020temporal}. We evaluate both prediction tasks and both datasets using \emph{Leave One User Out} as described in \cite{gao2014jhu}. 

Both trajectory and state predictions are performed in a multi-step manner for up to 2-second into the future ($max(T_{pred})=20$ for 10Hz data streams). The model performances with respect to the number of prediction time-steps are discussed in the next section. For each prediction time-step, the evaluation metrics are based only on the prediction at that time-step, without accounting for the previous prediction steps. To describe the representative performance of our model, we select a 1 second prediction time-step for Fig. \ref{fig:Figure6}, Table II, and Table III. 

\section{Results and Discussions} \label{sec:results}

\begin{figure}[t]
    \centering
    \medskip
    \includegraphics[width=10cm]{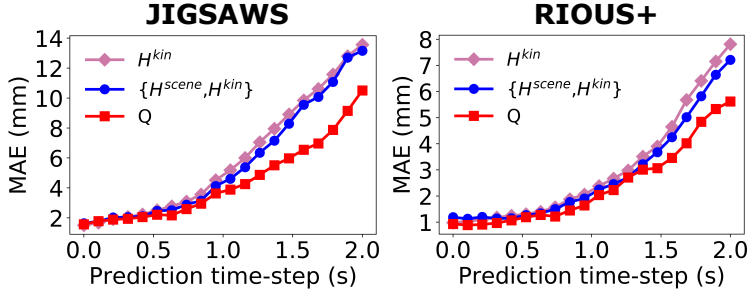}
    \caption{Model performance comparisons when different features are included for end-effector trajectory prediction for various prediction time-steps. The model was constructed with only kinematics features ($\pmb{H}^{kin}$), scene and kinematics features (\{$\pmb{H}^{scene}$, $\pmb{H}^{kin}$\}), and scene, RoI and kinematics features ($\pmb{Q}$). $mean(MAE_{d_1}+MAE_{d_2})$ and $MAE_d$ were plotted for JIGSAWS and RIOUS+, respectively.}
    \label{fig:Figure4}
\end{figure}

\begin{figure}[t]
    \centering
    \includegraphics[width=10cm]{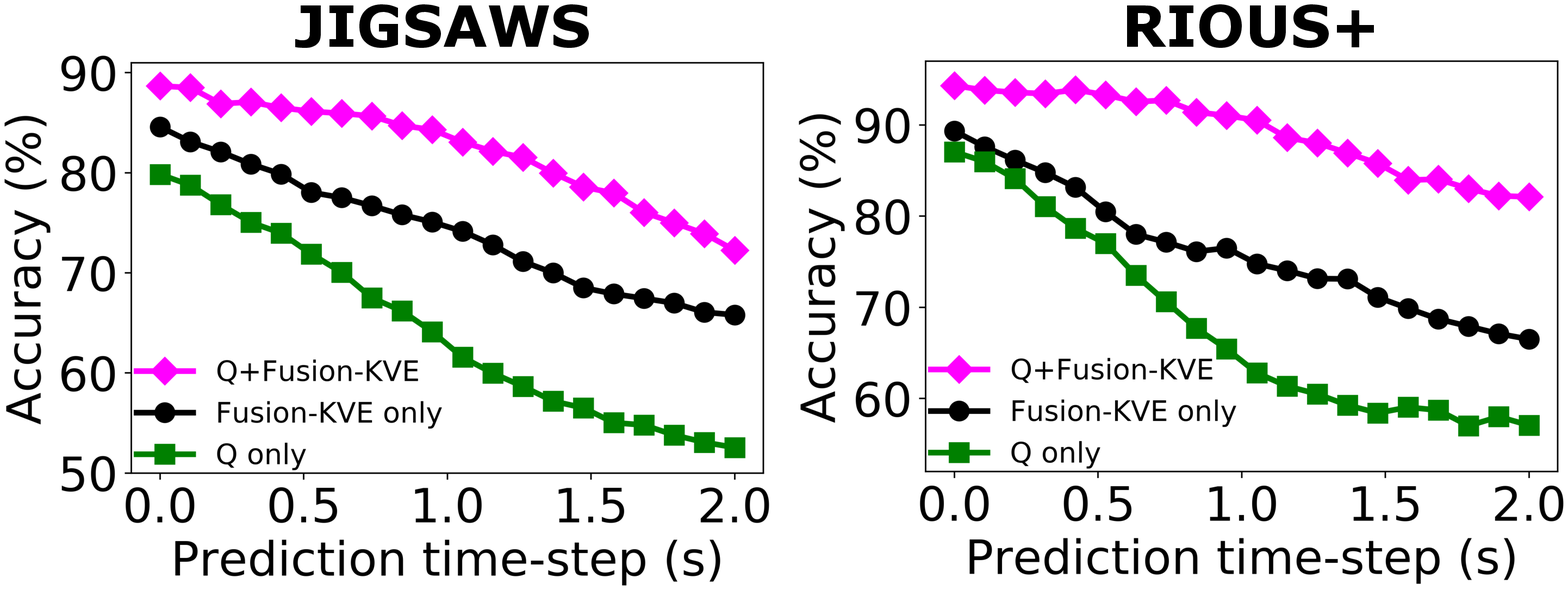}
    \caption{Model performance comparisons when different features are included for surgical state prediction. The state prediction decoder was trained with only feature tensor ($\pmb{Q}$ only), only historic state sequence (Fusion-KVE only), and both ($\pmb{Q}$+Fusion-KVE).}
    \label{fig:Figure5}
\end{figure}

\begin{figure*}
    \centering
    \medskip
    \includegraphics[width=\textwidth]{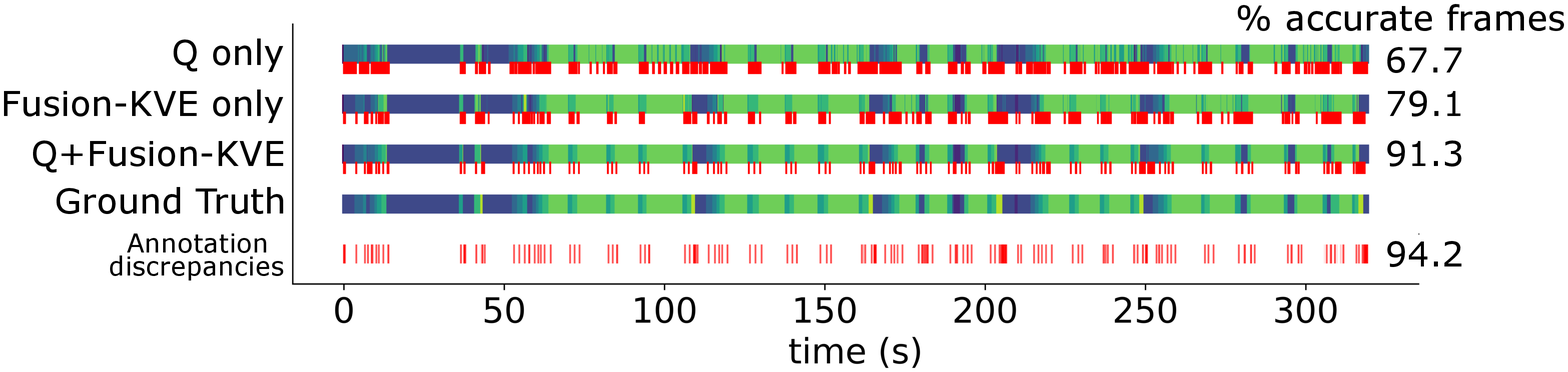}
    \caption{Sample 1-second ultrasound imaging state prediction results using only the feature tensor (\pmb{Q} only), only the historic state sequence (Fusion-KVE only), and both (\pmb{Q}+Fusion-KVE). Each block bar contains the state prediction results when $T_{pred}=10$ (top), and the discrepancies between the prediction results and the GT shown in red. The Annotation discrepancies row (bottom) shows the locations of frames where multiple annotators used different state labels, with the mean matching rate of 94.2\% among annotators.}
    \label{fig:Figure6}
\end{figure*}

Tables II and III summarize \emph{daVinciNet}'s end-effector path prediction and surgical state prediction performance on the JIGSAWS suturing and RIOUS+ datasets when the prediction time-step is 1 second ($T_{pred}=10$). Fig. \ref{fig:Figure4} and Fig. \ref{fig:Figure5} illustrate how our model performance changes at various prediction time-steps. Both tables and figures also include the performance of ablated versions of our model as compared to their performances on the full model. Fig. \ref{fig:Figure6} shows a sample sequence of ultrasound imaging task state prediction when $T_{pred}=10$ using our state prediction model as well as ablated versions of it.

Table II compares the differences in end-effector path prediction accuracy when \emph{daVinciNet} uses only kinematics features $\pmb{H}^{kin}$, scene and kinematics features \{$\pmb{H}^{scene}$, $\pmb{H}^{kin}$\}, and a full feature tensor $\pmb{Q}=\{\pmb{H}^{scene}, \pmb{H}^{RoI}, \pmb{H}^{kin}\}$. The accuracy of end-effector trajectory prediction in the endoscopic reference frame was evaluated, along with the end-effector distance $d=\sqrt{x^2+y^2+z^2}$ from the origin (camera tip).  \emph{daVinciNet} predictions based on all data streams consistently achieve up to 20\% better performance. Clearly, vision features contribute to better prediction of the end-effector trajectory in the endoscopic reference frame. Many instrument movements have advanced visual cues. E.g., suture pulling usually occurs after the needle tip has appeared on the suturing pad or tissue. Visual features, such as the distance between the end-effector and nearby tissue, also help in predicting trajectory changes. Therefore, including visual features, especially RoI information around the end-effectors, is helpful in end-effector trajectory prediction. Table III investigates the surgical state prediction accuracy with only $\pmb{Q}$, only state estimation results, or both as input features. The significant improvement in state prediction accuracy by incorporating both data sources supports our model design. As mentioned in the previous section,  \emph{daVinciNet} does not have access to ground truth state sequence in real-time prediction. The high prediction accuracy using an estimated state sequence shows the robustness of our model in real-time state prediction.

Fig. \ref{fig:Figure4} shows how end-effector trajectory prediction accuracy changes with increasing prediction time-step. We compared trajectory prediction MAE when the feature tensor includes only $\pmb{H}^{kin}$, \{$\pmb{H}^{scene}$, $\pmb{H}^{kin}$\}, or all three types of features, $\pmb{Q}$. For JIGSAWS, the MAE of the left ($d_1$) and right ($d_2$) instruments are averaged. The use of RoI visual features consistently decreases the trajectory prediction MAE, especially at large prediction steps. This observation reaffirms our discussion earlier that the visual features and advanced cues concentrated in RoIs were detected by the tool-scene encoder and contributed to end-effector trajectory prediction. 

\begin{table}[]
\centering
\medskip
\normalsize
\begin{tabular}{|c|c|c|}
\hline
Input data             & JIGSAWS Suturing (\%) & RIOUS+ (\%)          \\ \hline
\textbf{Q} only                 & 64.11            & 65.44          \\
Fusion-KVE only        & 75.08            & 76.5           \\
\textbf{Q}+Fusion-KVE  & \textbf{84.3}    & \textbf{91.02} \\ \hline
\end{tabular}
\caption*{Table III: Surgical State Prediction performance when predicting one second ahead ($T_{pred}=10$). The prediction performances are compared when the state prediction decoder uses only the feature tensor ($\pmb{Q}$ only), only the historic state sequence (Fusion-KVE only), and both ($\pmb{Q}$+Fusion-KVE).}
\end{table}

Fig. \ref{fig:Figure5} shows the progress of surgical state prediction accuracy as prediction time-step increases. For surgical state prediction, we compared performance when the state prediction decoder is based only on $\pmb{Q}$, only on the Fusion-KVE state sequence results, and both. The ablated prediction models were constructed with the $\tilde{\pmb{q}}$ or $\pmb{y}$ term omitted from Eq. (\ref{eq:beta})-(\ref{eq:d_t}), respectively. When only feature tensor $\pmb{Q}$ was available to the state prediction decoder, state prediction accuracy is constantly the lowest and exponentially decreases as the prediction time-step increase. Due to the absence of historic target series, the state prediction model can only rely on predicted state sequences from previous time-steps, which causes previous prediction errors to affect the next state prediction with little correction. When the state prediction decoder can use historic state estimation sequences from Fusion-KVE, prediction accuracy is improved, especially for large prediction steps, since the target series provide corrections to previous prediction errors. When the decoder receives only the target series with no additional feature series, the surgical state predictions are performed without visual or kinematic cues that indicate state changes; therefore, the improvement in prediction accuracy is limited. By incorporating both $\pmb{Q}$ and Fusion-KVE, advanced cues from visual and kinematics features are used to forecast state changes, and the historic state sequence provides corrections to prediction errors. \emph{daVinciNet} achieves the highest state prediction accuracy that does not show significant deterioration as prediction time-step increases.

The sample sequence of ultrasound imaging state prediction results in Fig. \ref{fig:Figure6} further supports our model architecture's inclusion of the feature tensor $\pmb{Q}$ and Fusion-KVE output for surgical state prediction. Prediction errors occur in blocks when only $\pmb{Q}$ is used for prediction, due to uncorrected errors from using the predicted state sequence from previous time-steps. A model using only Fusion-KVE shows fewer errors in consecutive time blocks; however, the missing feature input leads to delayed responses to real world state changes, or even missing states with relatively shorter duration. A \emph{daVinciNet} model that incorporates both $\pmb{Q}$ and Fusion-KVE significantly improves state prediction accuracy, with the remaining errors located mostly around state transitions. 

Additionally, since the temporal annotations of surgical states were done manually by humans, we investigated the annotation variance in the ground truth state sequence introduced by human annotators. Five users were asked to annotate the sample sequence in Fig. \ref{fig:Figure6} frame-by-frame with states in Table I. The discrepancies among annotations are shown in the bottom row of Fig. \ref{fig:Figure6}, with an average matching rate of 94.2\% among annotators. Even human annotators cannot agree perfectly on a state sequence: their discrepancies occur mostly at state transitions, which is expected since the transition from one state to another in a surgical subtask is not abrupt, but gradual. Hence, annotators may identify different video frames as state transitions. Hence,  \emph{daVinciNet} state prediction errors can be partially attributed to human annotation errors in identifying the exact state transition times. Thus, \emph{daVinciNet}'s robustness is further established by its high state prediction accuracy even in the presence of noise in the ground truth data.

\section{Conclusions and Future Work}

This paper focused on real-time prediction of variables that are crucial to RAS, including instrument end-effector trajectories in endoscopic viewing frames and discrete surgical states. We proposed the \emph{daVinciNet}: a unified end-to-end joint prediction model that uses synchronized sequences of robot kinematics, endoscopic vision, and system events data as input to predict instrument trajectories and surgical states. Our model achieves accurate predictions of the end-effector path, with distance error as low as 1.64mm and MAPE of 1.72\% when predicting the end-effector location 1-second in the future. The surgical state estimation accuracy achieved by the \emph{daVinciNet} is up to 91.02\%, and compares well with human annotator accuracy of 94.2\%. By accurately predicting the end-effector trajectory and surgical states in datasets with various experimental settings and robot motions, the \emph{daVinciNet} proves its robustness in realistic RAS tasks.

We further illustrated the necessity and advantages of including multiple data sources for joint prediction tasks by comparing the performance of our full model against ablated versions. Improved performance arises, for instance, because many instrument movements have visual features and advanced cues that can be captured by \emph{daVinciNet}'s endoscopic vision feature module. Including a full feature tensor with the historic state sequence also significantly improves accuracy in surgical state prediction, compared to only using one type of input. Richer information regarding surgical subtasks can be extracted from multiple encoders, which leads to more accurate predictions. We also showed the sizeable contribution of RoI visual features to performance. \emph{daVinciNet} incorporates a silhouette-based instrument tracking algorithm to identify the RoIs in endoscopic vision and our Fusion-KVE state estimation model \cite{qin2020temporal} to obtain the historic surgical state sequence. The applications of existing surgical scene-understanding models allow us to achieve better performances in prediction. 

To further improve prediction performance, a possible next step would be to incorporate semantic segmentation of endoscopic images. While extracting both global and RoI visual features provides \emph{daVinciNet} with rich information and advanced cues for better predictions, the backgrounds in real-world RAS video is complicated and diverse. A semantic segmentation model maps each pixel of the endoscopic vision to a semantic class and therefore reduces the environmental noise. We also plan to apply the \emph{daVinciNet} to RAS applications such as multi-agent systems with shared control or supervised autonomous surgical subtasks.


%%%%%%%%%%%%%%%%%%%%%%%%%%%%%%%%%%%%%%%%%%%%%%%%%%%%%%%%%%%%%%%%%%%%%%%%%%%%%%%%

%%%%%%%%%%%%%%%%%%%%%%%%%%%%%%%%%%%%%%%%%%%%%%%%%%%%%%%%%%%%%%%%%%%%%%%%%%%%%%%%

%%%%%%%%%%%%%%%%%%%%%%%%%%%%%%%%%%%%%%%%%%%%%%%%%%%%%%%%%%%%%%%%%%%%%%%%%%%%%%%%
\section*{ACKNOWLEDGMENT}
This work was funded by Intuitive Surgical, Inc. We would like to thank Dr. Azad Shademan and Dr. A. Jonathan McLeod for their support of this research.

%%%%%%%%%%%%%%%%%%%%%%%%%%%%%%%%%%%%%%%%%%%%%%%%%%%%%%%%%%%%%%%%%%%%%%%%%%%%%%%%

\bibliographystyle{IEEEtran}
\bibliography{IEEEabrv,IEEEexample}

\end{document}